%% file: main.tex
\documentclass{bmvc2k}

\usepackage{booktabs}
\usepackage{multirow}
\usepackage{mathdots}
\usepackage{algorithm}
\usepackage{algpseudocode}

\title{Information Removal at the bottleneck\\in Deep Neural Networks}

\addauthor{Enzo Tartaglione}{enzo.tartaglione@telecom-paris.fr}{1}

\addinstitution{
 LTCI, T\'el\'ecom Paris,\\Institut Polytechnique de Paris\\
}

\runninghead{Tartaglione}{Information Removal in Deep Neural Networks}

\def\etal{\emph{et al}\bmvaOneDot}

\def\f{\mathcal{F}}
\def\g{\mathcal{G}}
\def\h{\mathcal{H}}
\def\i{\mathcal{I}}
\def\l{\mathcal{L}}
\def\y{\hat{y}}
\def\v{\hat{v}}
\begin{document}

\maketitle

\begin{abstract}
Deep learning models are nowadays broadly deployed to solve an incredibly large variety of tasks. Commonly, leveraging over the availability of ``big data'', deep neural networks are trained as black-boxes, minimizing an objective function at its output. This however does not allow control over the propagation of some specific features through the model, like gender or race, for solving some an uncorrelated task. This raises issues either in the privacy domain (considering the propagation of unwanted information) and of bias (considering that these features are potentially used to solve the given task). 

In this work we propose IRENE, a method to achieve \underline{i}nformation \underline{re}moval at the bottleneck of deep \underline{ne}ural networks, which explicitly minimizes the estimated mutual information between the features to be kept ``private'' and the target. Experiments on a synthetic dataset and on CelebA validate the effectiveness of the proposed approach, and open the road towards the development of approaches guaranteeing information removal in deep neural networks.
\end{abstract}


\input{sections/1_intro.tex}
\input{sections/2_sota.tex}
\input{sections/3_method.tex}
\input{sections/4_results.tex}
\input{sections/5_conclusion.tex}

\bibliography{main}
\end{document}

%% file: sections/1_intro.tex
\section{Introduction}
\begin{figure}[t]
    \centering
    \includegraphics[width=0.8\columnwidth]{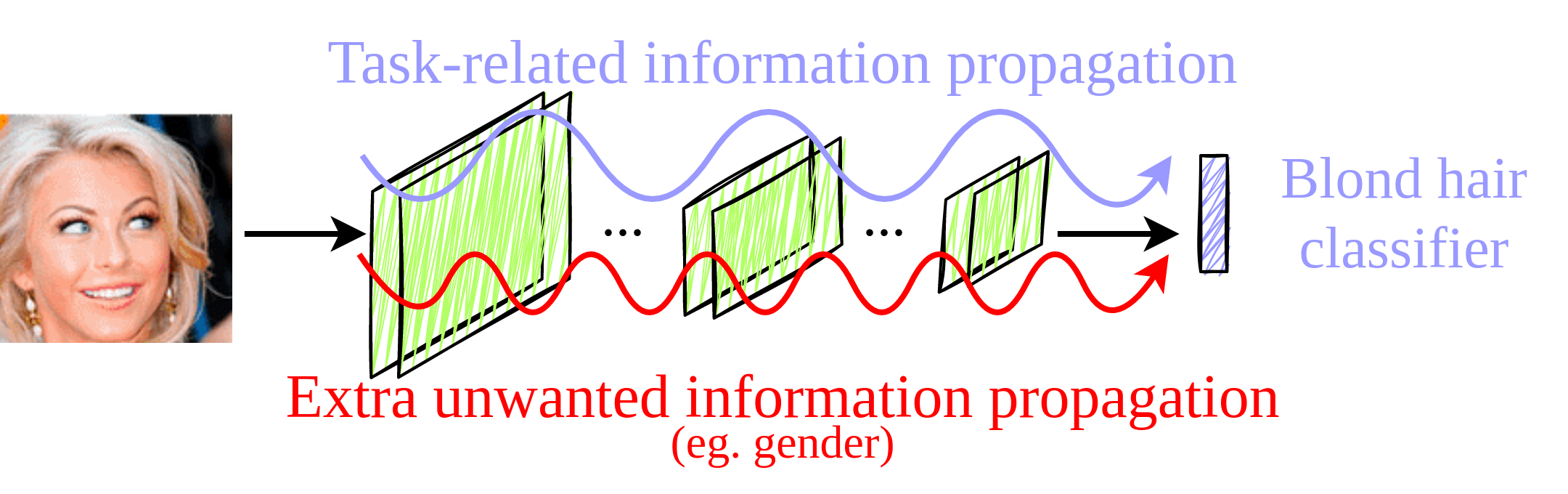}
    \caption{From the latent space of ANN models there might be information leakage, allowing an attacker to recover sensitive/private data. Our goal is to prevent this, hiding the private information.}
    \label{fig:teaser}
\end{figure}
Currently, a significantly large portion of problems is being solved through the deployment of deep learning models, considered by most the ``universal problem solving tool''~\cite{sonoda2017neural}. For instance, these are being deployed in high-stakes applications, ranging from candidate job hiring to facial recognition systems. It is a known problem that deep models, when trained, take advantage of ``spurious'' correlations from their training data, which lead to significant performance variance across sub-populations, sometimes across sensitive attributes like race and gender. This causes what is known, in the literature, as \emph{bias} of the deep model.

Learning these spurious correlations has several effects, of which the most evident one is poor performance on under-represented dataset sub-populations and out-of-distribution test data~\cite{krishnakumar2021udis}. Finding a solution to the problem of biases in the deep models is currently a topic of broad interest from the community~\cite{Sudhakar2021MitigatingBI, Henriksen2021BiasFR}. Many works have tried to address this problem, ranging from addressing the problem to transformers~\cite{Sudhakar2021MitigatingBI} to unsupervised scenarios~\cite{krishnakumar2021udis}. A large part of approaches intrinsically propose a re-weighting over the biased information, why not to entirely remove it?

A lot of interest around debiasing is indeed devoted because it would be irresponsible and unethical to promote algorithms that could possibly create damages or lean towards discrimination of people whose data are used by the Artificial Intelligence (AI)~\cite{Osia2020AHD, Li2021EverybodyIU}. The European Union has already obtained a role of legal influencer as far as the discipline of data protection is concerned by drafting the General Data Protection Regulation (GDPR)~\cite{art1} and because it has started since 2018 in trying to create the conditions to regulate the AI. Towards this end, re-weighting the information in the learning process might not be sufficient to guarantee complete protection towards the propagation (and for instance, the use) of some sensitive, or discriminatory information, by the AI model~\cite{barbano2021bridging}. Figure~\ref{fig:teaser} depicts the scenario in which, besides the task-related information (in this case, the color of the hairs), the information of the gender is also propagated. This behavior of deep models poses issues from ethical, security and legal perspectives.

In this work we propose IRENE, a method to remove specific information at the bottleneck of deep neural networks. This method relies on the estimation of the information we desire to maintain ``private'', with the employment of an auxiliary classifier. This enables the estimation of the information (at the bottleneck) we wish to remove. Then, we achieve the information removal through the minimization of a differentiable proxy of the mutual information. To the best of our knowledge, IRENE is the first work proposing an information removal at the bottleneck of deep neural networks.

The rest of the paper is organized as follows. In Section~\ref{sec:related} an overview on the close, related literature is provided; section~\ref{sec:IRENE} presents IRENE, then Section~\ref{sec:expers} shows and discusses the experiments conducted. Finally, Section~\ref{sec:conclusion} draws the conclusions.

%% file: sections/2_sota.tex
\section{Related works}
\label{sec:related}

In this section we discuss the literature related to information removal in deep learning. We can group the related literature in two large families: privacy preservation approaches and debiasing techniques.\\ 

\textbf{Privacy preservation.} 
Very recently, thanks to the increase of computational capabilities, many works have been proposed on privacy-preserving in computational frameworks. A work by Dwork \etal~\cite{dwork2016calibrating} studied how much noise is required to guarantee ``differential privacy''~\cite{dwork2009differential} from data. Duchi \etal~\cite{duchi2014privacy} formalized convergence boundaries for training and the trade-off between privacy guarantees and the utility of the resulting statistical estimators. 
This knowledge has been also recently applied to deep learning frameworks, with a work by Abadi \etal~\cite{abadi2016deep}, by introducing some tuned noise in the update rule. A different approach to preserve data privacy is the so-called ``federated learning''. In general, private datasets are held by the proprietary of the data, who is directly training a local neural network model. Now, the parameters of the models are sent to a master node, who is then propagating to all the private computational nodes the general configuration of the parameters. This approach has been proposed by Shokri and Shmatikov~\cite{shokri2015privacy}, and allows parallel and private computation. However, it does not take into account any ethical bias, like gender or race: what it guarantees is that the original data are not directly shared, but some sensible information actually are.

\textbf{Debiasing.} It is known that datasets are typically affected by biases. In their work, Torralba~and~Efros~\cite{torralba2011unbiased} showed how biases affect some of the most commonly used datasets, drawing considerations on the generalization performance and classification capability of the trained ANN models. Following-up on a similar idea, the effectiveness of content-style disentanglement has been explored, besides an estimation of bias content content, by Liu \etal~\cite{Liu2021MeasuringTB}. Working at the dataset level is in general a critical aspect, and greatly helps in understanding the data and its structure~\cite{Cubuk_2019_CVPR}. 
Some works suggest the use of GANs to entirely clean-up the dataset with the aim of providing fairness~\cite{xu2018fairgan, sattigeri2018fairness}, others like Mandras~\etal~\cite{madras2018learning} insert a GAN in the middle of the architecture to clean-up the internal representation of data. In general, training such an architecture is a very delicate and complex process, and it does not provide explicit fairness at inference time, as generative adversarial networks are used to generate training data. Another interesting possibility has been proposed by Kim~\etal, with the use adversarial learning and gradient inversion to eliminate the information related to the biases in the model~\cite{Kim_2019_CVPR}. 
Bahng~\etal~\cite{bahng2019rebias} develop an ensembling-based technique, called \emph{ReBias}: this consists in solving a min-max problem where the target is to promote the independence between the network prediction and all biased predictions. Identifying the ``known unknowns''~\cite{attenberg2015beat} and optimizing on those (using a neural networks ensemble) is the approach proposed by Nam~\etal~\cite{nam2020learning}, or with unsupervised surface exploration, by Khrisnakumar~\etal~\cite{Krishnakumar2021UDISUD}. A similar approach is followed by Clark~\etal in their LearnedMixin~\cite{ClarkYZ19}. The exploration of the embedding space, looking for biases and addressing the bias mitigation problem and discouraging the optimization directions which favor the classifier to be biased, is proposed by Thong~\etal~\cite{Thong2021FeatureAL}. The inclusion of a regularization term, addressing a similar concern but with no memory overhead, has also been recently proposed by Tartaglione~\etal~\cite{tartaglione2021end}. 

Considering the specific problem IRENE addresses, despite having similar objectives to privacy preservation approaches, the setup of the problem makes IRENE closer to debiasing approaches, despite its different goal. A recent work by Song~\etal~\cite{song2017machine} reported that, using standard training strategies to train some state-of-the-art models, allows information not relevant to the learning task to be stored inside the network. Such a behavior is possible because of the typically oversized ANNs trained to solve a task~\cite{tartaglione2019take}. In their experiments, Song~\etal show how accurately they can recover some non-directly related to training information, showing the potential, unwanted information propagation. In the next section IRENE will be presented and discussed.

%% file: sections/3_method.tex
\section{IRENE: information removal at the bottleneck}
\label{sec:IRENE}
In this section we are going to introduce IRENE, our method to remove information at the bottleneck in deep neural networks. As we will see, for a given learning task, there can naturally be some mutual information between the private information ($\v$) and the target ($\y$). Typical debiasing approaches mitigate, or rather, balance these two, towards an improvement of the performance on some target task. In our case, we target the pure minimization of such a term. Here follows a general presentation of the learning framework and of the proposed approach.

\subsection{Private features and target features}

Let us train a given deep neural network such that, given an input $x$, it produces an output $y$: the vanilla training ambition is to make it as close as possible to the ground truth $\hat{y}$. Towards this end, a loss function $\mathcal{L}(y,\hat{y})$ is minimized (eventually, besides some additional regularization constraint). Hence, the only control over the information being learned lies in the ground truth label $\hat{y}$. 

Let us define here, for every input sample $x$, a companion ground truth label $\hat{v}$, marking a different information from the task-related one: this one we wish \emph{not} to be propagated in the model, or rather, not to be propagated from a certain layer (we define this \emph{bottleneck layer}) onward. Naturally, there exists some correlation between the $i$-th task-related ground truth $\hat{y}$ and the $j$-th $\hat{v}$, which can be modeled by the joint probability $p_{\hat{y}\hat{v}}$. Having this, we know the mutual information $\mathcal{I}(\hat{y}, \hat{v})$ being
\begin{equation}
    \mathcal{I}(\hat{y}, \hat{v}) = \sum_i \sum_j p_{\hat{y}\hat{v}}(i,j) \log\left( \frac{p_{\hat{y}\hat{v}}(i, j)}{p_{\hat{y}}(i) p_{\hat{v}}(j)}\right).
\end{equation}

In the case of a perfect learner (ie. a model making no errors on the training set), since $y^\mu = \hat{y}^\mu~\forall \mu$ (where $\mu$ is then input sample index), necessarily we will have $\mathcal{I}(y, \hat{v}) = \mathcal{I}(\hat{y}, \hat{v})$. However, from the debiasing literature, in the cases when there exist certain $p_{\hat{y}\hat{v}}(i,j) \gg p_{\hat{y}\hat{v}}(i,k)$, the learning is imperfect as there are attractors in the learning process, making $\mathcal{I}(y, \hat{v}) > \mathcal{I}(\hat{y}, \hat{v})$. In such a context, the debiasing literature leverages over such a gap in order to extract the information over $p_{\hat{y}\hat{v}}$ imbalances, correcting them. The purpose of debiasing algorithms; however, is not to achieve a ``perfect learner'' on the (biased) training set, but to improve the performance on the (unbiased) validation set, where $p_{\hat{y}\hat{v}}(i,j) \approx K \forall i,j$ (hence, uniformly distributed).

Let us divide our model into an encoder $\f$ and a classifier $\g$: we name the output of the encoder \emph{bottleneck} $z$, which is also the input of $\g$. In order to improve the pure performance of the classifier model, it will be a necessary condition to have $\mathcal{I}(\hat{y}, \hat{v}) = \mathcal{I}(y, \hat{v})$ but this is not sufficient to guarantee that $\mathcal{I}(\hat{y}, \hat{v}) = \mathcal{I}(z, \hat{v})$. Indeed, debiasing methods target a balanced extraction (and eventually, balanced use) for $\hat{v}$: the removal/decorrelation with respect to the target task is not necessarily addressed to achieve their purposes. The task we want to achieve is to explicitly minimize $\mathcal{I}(z, \hat{v})$: here follows our method designed to address such a task.

\subsection{How to remove information at the bottleneck}
\label{sec:methodremove}
\begin{figure}
    \centering
    \includegraphics[trim={90 30 450 0},clip, width=1.0\columnwidth]{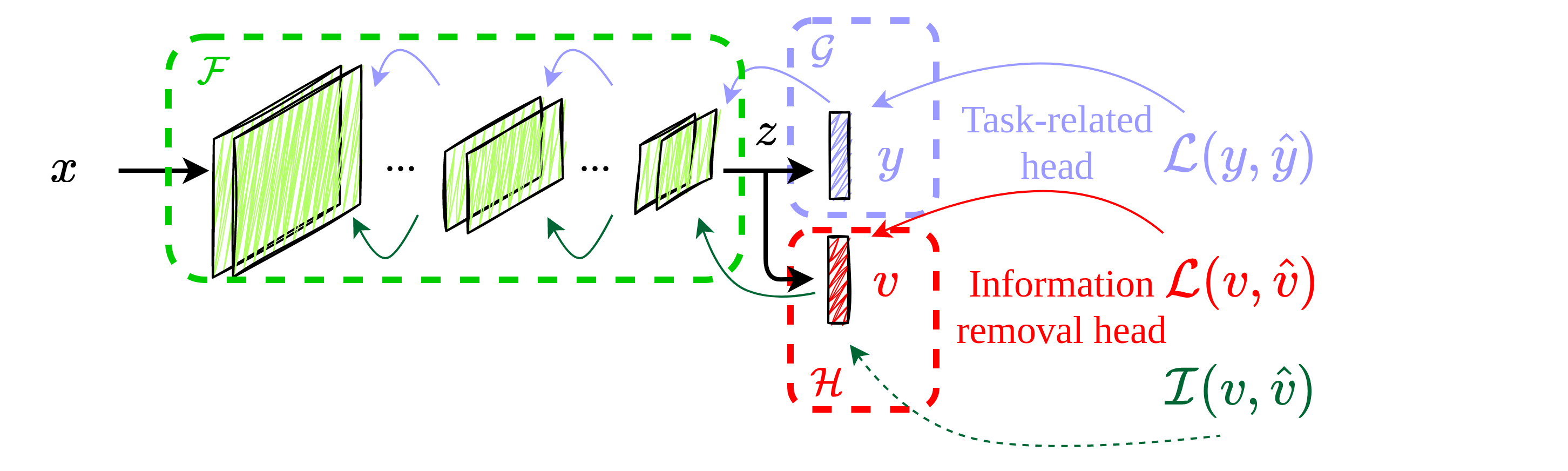}
    \caption{Sketch of IRENE. Backward lines display the back-propagation of the single contributions, dashed lines imply non-computation of the gradients for the target layer.}
    \label{fig:backprop mechanism}
\end{figure}

\begin{algorithm}[ht]
\caption{Training with IRENE.}
\label{alg:irene}
\begin{algorithmic}[1]
\Procedure{One\_iteration\_IRENE($x$, $\hat{y}$, $\hat{v}$, $\theta_{\f}$, $\theta_{\g}$, $\theta_{\h}$)}{}
\State $z \gets \f(x, \theta_\f)$
\State $y \gets \g(z, \theta_\g)$
\State $v \gets \h(z, \theta_\h)$
\State $\text{grad}(\theta_\g) \gets \text{backward}(\alpha \mathcal{L}(y, \hat{y}))$
\State $\text{grad}(\theta_\h) \gets \text{backward}(\mathcal{L}(v, \hat{v}))$
\State $\text{grad}(\theta_\f) \gets \text{backward}(\alpha \mathcal{L}(y, \hat{y}) + \gamma \mathcal{I}(v, \hat{v}))$
\State Update with chosen optimizer
\EndProcedure
\end{algorithmic}
\end{algorithm}

Directly addressing the problem of minimizing $\mathcal{I}(z, \hat{v})$ is either computationally unfeasible or it requires to impose constraints on the architecture of the model itself. We choose to follow neither of these two roads, but instead to distill from $z$ how much information related from $\hat{v}$ is filtering. Towards this end, we need to train, besides the task-related head $\g(z,\theta_{\g})$, another head $\h(z,\theta_{\h})$, which is trained to extract $\hat{v}$ from $z$, where $\theta_{\g}$ and $\theta_{\h}$ are the parameters associated to $\g$ and $\h$, respectively. 

The output of $\h$ is the prediction over the feature willing to maintain private, thanks to it is possible to estimate how much information it is possible to extract at the bottleneck $z$ of the model through $\mathcal{I}(v, \hat{v})$. The overall gradient computation strategy for one iteration is presented in Algorithm~\ref{alg:irene}. After obtaining $y$ (line 3) and $v$ (line 4) from forward propagation, we can compute the three quantities of interest (to be minimized), namely:
\begin{itemize}
    \item The loss $\l(y, \hat{y})$, to be minimized in order to train the model to learn the target task. This term is scaled by a positive hyper-parameter $\alpha$ and back-propagated through $\f(x, \theta_\f)$ and $\g(z, \theta_\g)$; hence, both the groups of parameters $\theta_\f$ (line~7) and $\theta_\g$ (line~5) will be updated in order to contribute towards the minimization of this term.
    \item The loss $\l(v, \hat{v})$, to be minimized in order to train the information removal head $\h$ to extract all the information about $\hat{v}$ from the bottleneck $z$. This term is back-propagated through $\h$ only (line~6): indeed, allowing the back-propagation also through $\f$ would favor the leakage of the information of $\v$ to $z$ countering our whole purpose. On the contrary, the purpose of $\h$ is indeed to act as an estimator over the extractable information, which will be crucial for the correct estimation of the next term.
    \item The mutual information $\i(v,\v)$, to be minimized on order to accomplish our purpose of erasing the information from the bottleneck $z$ of the model. This term is scaled by a positive hyper-parameter $\gamma$, back-propagated through $\h$ and $\f$, but the gradients computed for $\f$ will be the only ones kept and maintained for the update step (line~7). It is of crucial importance not to use in the update step the gradient values in $\h$ as it will be in contrast with the learning problem as in the previous point (extracting the information about $\hat{v}$ from the bottleneck $z$). 
\end{itemize}
A graphical representation of the working principles for IRENE is also visualized in Figure~\ref{fig:backprop mechanism}, where the colored arrows represent the back-propagation for the individual terms to be minimized, and the dashed arrow represents back-propagation without gradient computation.

\subsection{Differentiable mutual information proxy}
In the previous sub-section we have presented our strategy to address the problem of minimizing the information of a target attribute $\v$ from the input $x$ at the bottleneck $z$. This includes as well the minimization of the mutual information $\i(\boldsymbol{v},\v)$. 

According to the definition of mutual information, in order to estimate the joint probability, we should necessarily extract the predicted label $\tilde{v}$ with $\tilde{v} = \text{argmax}(v_i)$, and from this we can compute the joint $p_{\tilde{v}\v}$. Unfortunately, this operation is non-differentiable; hence, we need to provide a smooth, differentiable operator in its place. Towards this end, similarly to what is done for minimizing the cross-entropy loss, we substitute it with the softmax $\sigma(\cdot)$ which assigns a normalized score to the outputs of $\h$, compatible with our setup. At such point, we can easily compute the differentiable proxy of the mutual information
\begin{equation}
    \mathcal{I}(v, \v) = \sum_i \sum_j p_{\sigma(v)\hat{v}}(i,j) \log\left( \frac{p_{\sigma(v)\hat{v}}(i, j)}{p_{\sigma(v)}(i) p_{\hat{v}}(j)}\right).
    \label{eq:MI}
\end{equation}
Minimizing \eqref{eq:MI} with the update strategy described in Section~\ref{sec:methodremove} drives the output of $\h$ towards maximum confusion, making $\sigma(v)_i\rightarrow\frac{1}{C}\forall i$, where $C$ is the number of classes for the information removal task. 

It is worth of noticing that, for our specific task, \eqref{eq:MI} can not be substituted with the maximization of the cross-entropy loss $\l(v,\v)$. Let us assume we have $C=2$: under the assumption that the training is completed with success, the classifier will always make the wrong prediction, which in the binary case results in the maximum mutual information. This effect needs to be taken into account when applying debiasing strategies to pure biased (or in our case, private) feature removal~\cite{bahng2019rebias,Thong2021FeatureAL,ClarkYZ19,Kim_2019_CVPR}. In the next section we present our empirical evaluation.

%% file: sections/4_results.tex
\section{Experiments}
\label{sec:expers}
In this section, we present the experiments we conducted to test IRENE on-the-field. We perform our experiments on two datasets: Biased-MNIST (synthetic) and CelebA (a celebrity faces dataset). In all the presented experiments we will train, besides the main model composed of $\f$ and $\g$, the auxiliary head $\h$, attempting to extract the information willing to remove at the bottleneck $z$ (which in our experiments we define as the layer before the output layer). Our training and inference algorithms are implemented in Python, using PyTorch~1.12 and a RTX3090~Ti NVIDIA GPU with 24GB of memory has been used for training and inference. \footnote{The source code will be made available at the conference's date, at the link \url{https://github.com/enzotarta/irene}.}

\subsection{Experiments on BiasedMNIST}
\begin{figure}
    \centering
    \includegraphics[width=\columnwidth]{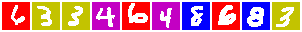}
    \caption{Example of images in the Biased-MNIST dataset. The background color correlates to a specific digit depending on the chosen value of $\rho$.}
    \label{fig:biased-mnist}
\end{figure}

\begin{figure}[t]
    \begin{tabular}{cc}
        \bmvaHangBox{\includegraphics[width=0.45\columnwidth]{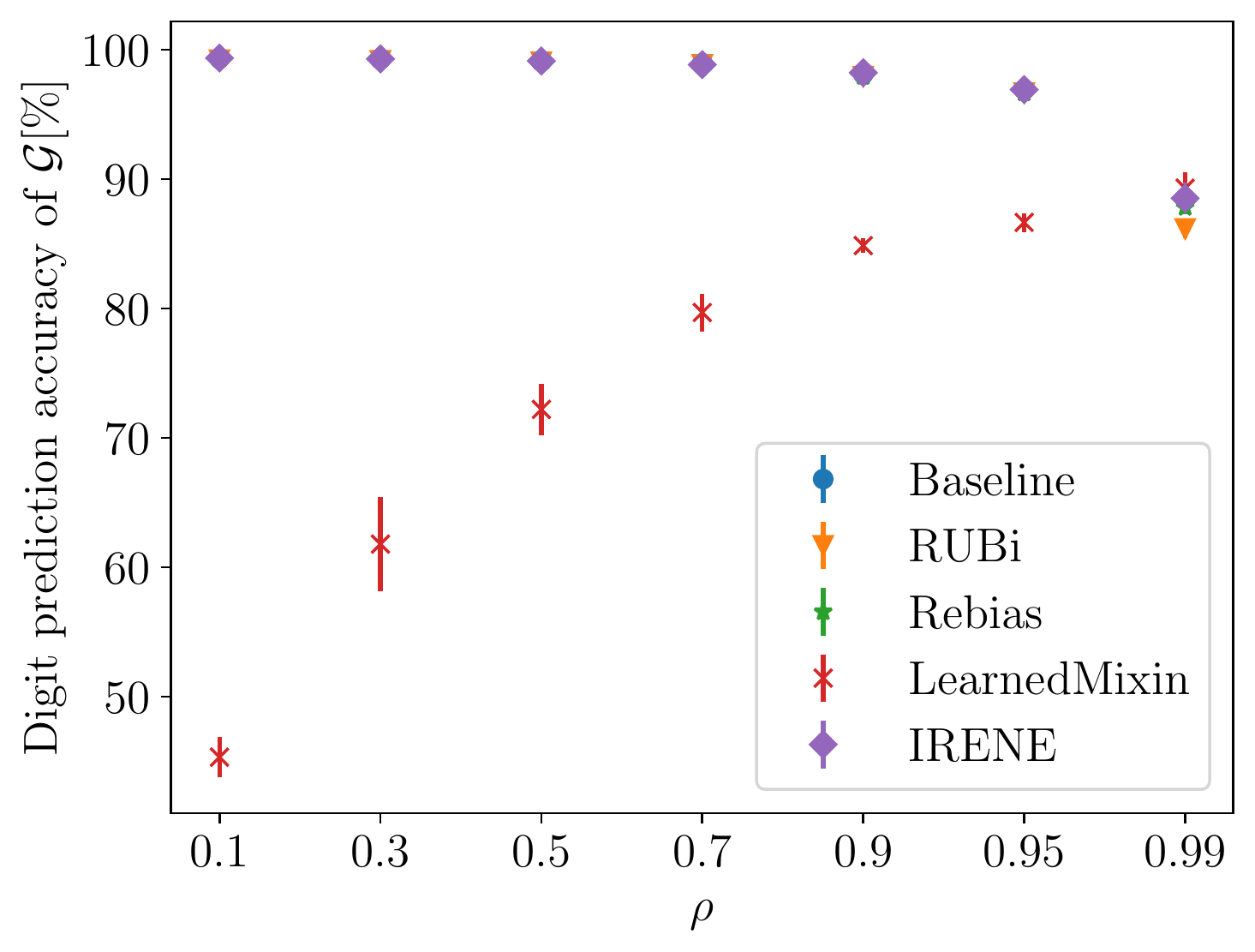}}&
        \bmvaHangBox{\includegraphics[width=0.45\columnwidth]{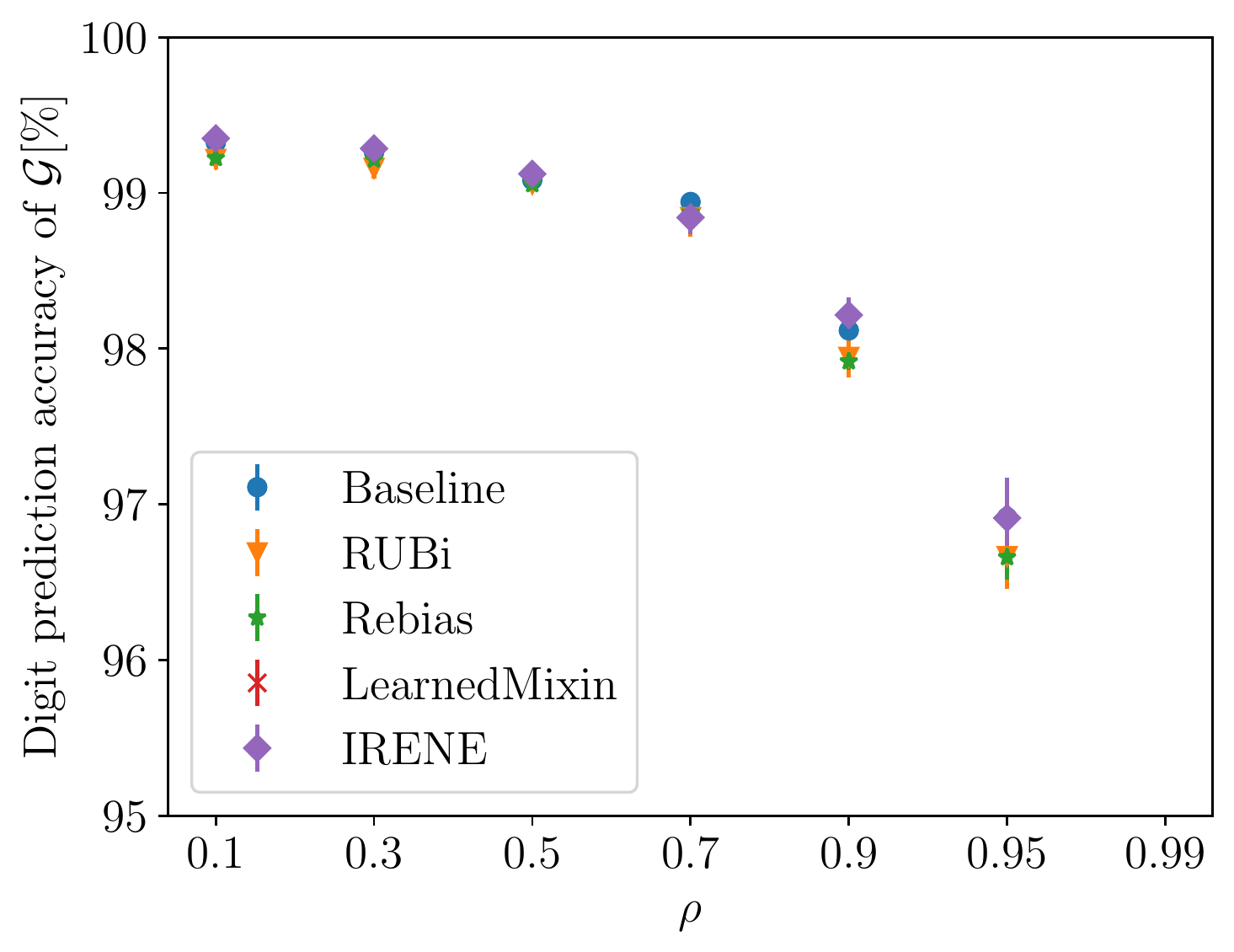}}\\
        (a)&(b)\\
        \bmvaHangBox{\includegraphics[width=0.45\columnwidth]{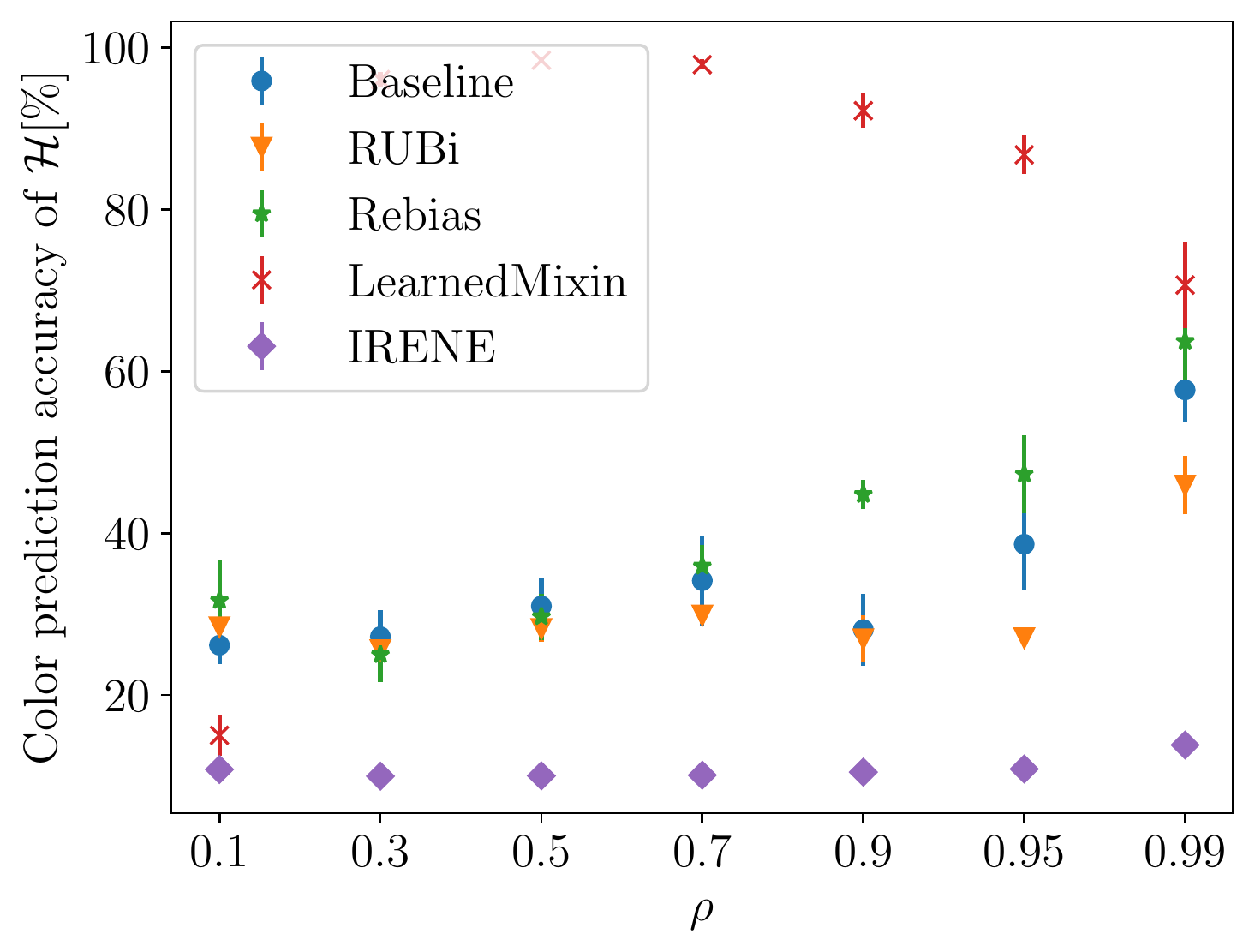}}&
        \bmvaHangBox{\includegraphics[width=0.45\columnwidth]{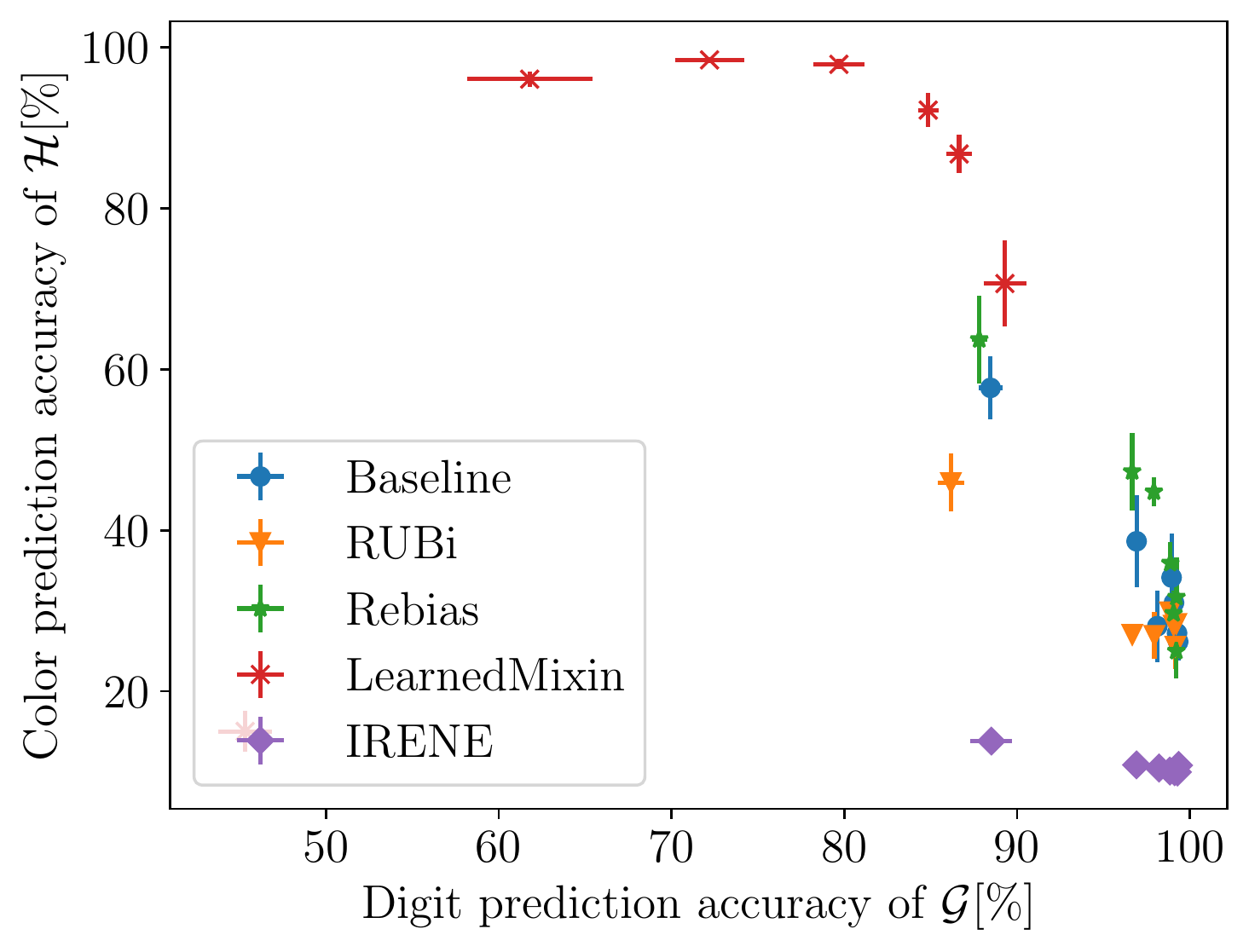}}\\
        (c)&(d)\\
    \end{tabular}
    \caption{Results on the BiasedMNIST dataset for different values of $\rho$. We report the performance of $\g$ (target task) as a function of $\rho$ (the higher, the better) (a) and its zooming (b), the performance of $\h$ (the lower, the better) (c) and the performance of $\h$ as a function of the performance achieved by $\g$ on the target task (lower right corner, the better) (d).}
    \label{fig:biasedmnist}
\end{figure}
As a first benchmark, we employ the synthetic dataset Biased-MNIST, recently proposed by Bahng~\etal~\cite{bahng2019rebias}. This dataset is built on top of the broadly-known MNIST dataset~\cite{lecun2010mnist}, adding a background color as displayed in Figure~\ref{fig:biased-mnist}. Its peculiarity relies in the possibility of correlating the background color to a specific digit with an hyper-parameter $\rho\in[0.1; 1.0]$ (where $0.1$ corresponds to uncorrelated and $1.0$ is total correlation), which can be freely modified. Despite this dataset is mainly employed by the debiasing literature with the purpose of improving the digit classification performance on an uncorrelated test set, we will use it to test IRENE and its effect on both the information removal task (information about the background color) and the digit classification performance (target task), measured on the test set.

In order to investigate over potential similarities and differences with some debiasing algorithms, we report as well results obtained with three debiasing techniques: RUBi~\cite{cadene2019rubi}, Rebias~\cite{bahng2019rebias} and LearnedMixin~\cite{ClarkYZ19}. 

Following \cite{bahng2019rebias}, we train a convolutional architecture consisting of four $7\times 7$ kernels, with ReLU activation and batchnorm layers between convolution and activation. We have tested $\rho \in \{0.1,0.3,0.5,0.7,0.9,0.95,0.99\}$ and averaged the results over 10 different runs. For all the experiments we have used SGD optimization for 80 epochs with learning rate $0.1$, decayed by a factor 0.1 at the milestones [40,60], weight decay $10^{-4}$ and batch size 100. For IRENE we have set $\alpha=0.5$ and $\gamma=0.5$.

Figure~\ref{fig:biasedmnist} reports the results of our experiments. Looking at the baseline result, we observe a physiological drop in the digit prediction accuracy as $\rho$ increases (and consequently the increment of the color prediction accuracy), which is a marker of the use of the background color for the prediction. Interestingly, for the uncorrelated training set scenario ($\rho=0.1$), the color prediction accuracy is above random guess (10\%), meaning that there is, in any case, information leakage. IRENE performs very well, maintaining the baseline performances on the target task for all the tested values of $\rho$ and maintaining the performance of $\h$ close to random guess. Similarly to what observed by Barbano~\etal~\cite{barbano2021bridging} (despite in a different range of values for $\rho$), debiasing strategies allow information leakage, which is however employed in order to improve the performance of $\g$. In these experiments we observe, for the debiasing algorithms, a performance close to the baseline as these algorithms focus on more extreme regimes (in these cases, higher $\rho$). 

\subsection{Experiments on CelebA}
\begin{table*}
    \caption{Results on the CelebA dataset. Here the gender is the information to erase.}
    \label{tab:celeba}
    \small
	\renewcommand{\arraystretch}{1.2}
	\centering
		\begin{tabular}{c c c c c c c }
        \toprule
        &  & \bf{Prediction accuracy}  & \bf{Gender prediction accuracy} \\
        \bf{Target}         &  \bf{Method}     & \bf{of $\boldsymbol{\g}$ (trained task)  }    &  \bf{of $\h$ (information to remove)} \\
                &&\bf{[\%]($\boldsymbol{\uparrow}$)}&\bf{[\%]($\downarrow$)}\\
        \midrule
            \multirow{4}{*}{Blond hair}
                &Baseline   &95.34$\pm$0.07 &84.32$\pm$2.76\\
                &IRENE ($\gamma=0.1$)&\bf{95.37}$\pm$0.10 &55.47$\pm$8.18\\
                &IRENE ($\gamma=0.5$)&95.28$\pm$0.09 &53.64$\pm$10.69\\
                &IRENE ($\gamma=1$)&95.24$\pm$0.29 &\bf{53.68}$\pm$10.71\\
            \midrule
            \multirow{4}{*}{Heavy makeup}
                &Baseline   &\bf{90.58}$\pm$0.14 &92.89$\pm$0.36\\
                &IRENE ($\gamma=0.1$)&90.32$\pm$0.97 &65.13$\pm$11.08\\
                &IRENE ($\gamma=0.5$)&85.66$\pm$2.80 &56.45$\pm$9.46\\
                &IRENE ($\gamma=1$)&83.31$\pm$3.41 &\bf{51.98}$\pm$9.56\\
        \midrule
            \multirow{4}{*}{Eyeglasses}
                &Baseline   &99.67$\pm$0.02 &69.51$\pm$4.33\\
                &IRENE ($\gamma=0.1$)&99.68$\pm$0.01 &64.08$\pm$1.08\\
                &IRENE ($\gamma=0.5$)&\bf{99.69}$\pm$0.02 &61.57$\pm$6.95\\
                &IRENE ($\gamma=1$)&99.68$\pm$0.01 &\bf{54.66}$\pm$12.58\\
        \bottomrule
	\end{tabular}
\end{table*}
The CelebA dataset~\cite{liu2015faceattributes} has been designed for face-recognition tasks, providing 40 attributes for every image. The dataset contains a total of 202.6k images and, following the official train-validation split, we obtain 162.7k images for training set, 19.9k images as validation set and 19.9k images for testing our models. For our training purposes, we use a ResNet-18 model backbone as $\f$. 

The training has been performed using SGD, with an initial learning rate of 0.1, decayed by a factor 10 after no improvement over the validation set loss has been detected for 10 consecutive epochs. The training stops when the learning rate drops below $10^{-3}$. We use batch size 100 with momentum of 0.9 and weight decay of $10^{-5}$. Images are here re-scaled to $224\times 224$ pixels. For IRENE, we use $\alpha=0.5$ for all the experiments while $\gamma\in\{0.1, 0.5,1\}$. 

Differently from the debiasing literature, we are here not interested in testing the performance of a model trained over an unbalanced dataset, but we are here interested in observing the impact of some specific features we wish to keep private over the target task. Towards this end, we perform training over a \emph{balanced} training set. We select here as classification target the recognition of the \emph{eyeglasses}, \emph{blond hair} and \emph{heavy makeup} attributes, while the gender is the information we wish to remove.

Results are reported in Table~\ref{tab:celeba} and are averaged over 10 seeds. Here we observe different behaviors depending on the target task. Intuitively, the eyeglasses task (identifying the wearing of lenses) should be the one more ``disentangled'' by the gender information, and indeed, even with high emphasis on the private feature removal (with the highest $\gamma$) the trained task performance remains close to the baseline. For the \emph{blond hair} task (identifying people with blond hairs), we observe that the performance on the target task remains close to the baseline but the information leakage on the gender is higher than for eyeglasses; however, even with a low $\gamma$, this information can be successfully removed. Finally, with \emph{heavy makeup} (identifying people with a heavy makeup), we observe the highest performance of $\h$ over all the tasks (in this case, even higher than $\g$) and removing this information worsens the performance on the target task. Evidently, the information of gender is crucial to identify different makeup styles. From these three cases we conclude that it is in general possible to remove information on a specific feature at the bottleneck of a deep model, but depending on the target task the performance might be impacted.


%% file: sections/5_conclusion.tex
\section{Conclusion and future work}
\label{sec:conclusion}
In this work we have proposed IRENE, a method to remove information at the bottleneck in deep neural networks. In particular, we train an auxiliary head classifier $\h$ which extracts the information on the ``private'' classes at the bottleneck $z$, and this is used to minimize directly at the bottleneck the mutual information between the ground truth of the information to remove and $z$. 

We have tested IRENE on a synthetic dataset, BiasedMNIST, on 7 different regimes, and on CelebA, a dataset of faces of celebrities, with three different target tasks. For all the experiments IRENE is able to remove the information being asked to, but depending on the specific target task the performance might be affected. Interestingly, IRENE performs similarly to the tested debiasing algorithms in terms of target accuracy while removing the information to keep private. This suggests that extending IRENE to debiasing applications is a promising future research direction.